# Order-of-Magnitude Influence Diagrams[*]

**Radu Marinescu**
IBM Research – Dublin
Mulhuddart, Dublin 15, Ireland
radu.marinescu@ie.ibm.com

**Nic Wilson**
Cork Constraint Computation Centre
University College Cork, Ireland
n.wilson@4c.ucc.ie

## Abstract

In this paper, we develop a qualitative theory of influence diagrams that can be used to model and solve sequential decision making tasks when only qualitative (or imprecise) information is available. Our approach is based on an order-of-magnitude approximation of both probabilities and utilities and allows for specifying partially ordered preferences via sets of utility values. We also propose a dedicated variable elimination algorithm that can be applied for solving order-of-magnitude influence diagrams.

## 1 INTRODUCTION

Influence diagrams have been widely used for the past three decades as a graphical model to formulate and solve decision problems under uncertainty. The standard formulation of an influence diagram consists of two types of information: *qualitative information* that defines the structure of the problem eg, the set of (discrete) chance variables describing the set of possible world configurations, the set of available decisions, as well as the dependencies between the variables, and *quantitative information* (also known as the parametric structure) that, together with the qualitative information, defines the model. The parametric structure is composed of the conditional probability distributions as well as the utility functions describing the decision maker's preferences. In general, the solution to an influence diagram depends on both types of information. Quite often, however, we may have precise knowledge of the qualitative information but only very rough (or imprecise) estimates of the quantitative parameters. In such cases, the standard solution techniques cannot be applied directly, unless the missing information is accounted for.

In this paper, we propose a qualitative theory for influence diagrams in which such partially specified sequential decision problems can be modeled and solved. In particular, we introduce the order-of-magnitude influence diagram model that uses an order-of-magnitude representation of the probabilities and utilities. The model allows the decision maker to specify partially ordered preferences via finite sets of utility values. In this case, there will typically not be a unique maximal value of the expected utility, but rather a set of them. To compute this set and also the corresponding decision policy we propose a dedicated variable elimination algorithm that performs efficient operations on sets of utility values. Numerical experiments on selected classes of influence diagrams show that as the quantitative information becomes more precise, the qualitative decision process becomes closer to the standard one.

The paper is organized as follows. Section 2 gives background on influence diagrams. In Section 3 we present the order-of-magnitude calculus as a representation framework for imprecise probabilities and utilities. Sections 4 and 5 describe the main operations over sets of order-of-magnitude values and introduce the order-of-magnitude influence diagram model. In Section 6 we present the results of our empirical evaluation. Section 7 overviews related work, while Section 8 provides concluding remarks.

## 2 INFLUENCE DIAGRAMS

An *influence diagram* is defined by a tuple $\langle \mathbf{X}, \mathbf{D}, \mathbf{U}, G \rangle$, where $\mathbf{X} = \{X_1, \ldots, X_n\}$ is a set of oval-shaped nodes labeled by the *chance variables* which specify the uncertain decision environment, $\mathbf{D} = \{D_1, \ldots, D_m\}$ is a set of rectangle-shaped nodes labeled by the *decision variables* which specify the possible decisions to be made in the domain, $\mathbf{U} = \{U_1, \ldots, U_r\}$ are diamond-shaped nodes labeled by the *utility functions* which represent the preferences of the decision maker, and $G$ is a directed acyclic graph containing all the nodes $\mathbf{X} \cup \mathbf{D} \cup \mathbf{U}$. As in belief networks, each chance variable $X_i \in \mathbf{X}$ is associated with a conditional probability table (CPT) $P_i = P(X_i | pa(X_i))$, where $pa(X_i) \subseteq \mathbf{X} \cup \mathbf{D} \setminus \{X_i\}$ are the parents of $X_i$ in $G$. Similarly, each decision variable $D_k \in \mathbf{D}$ has a parent set $pa(D_k) \subseteq \mathbf{X} \cup \mathbf{D} \setminus \{D_k\}$ in $G$, denoting the variables whose values will be known at the time of the decision and may affect directly the decision. *Non-forgetting* is typically assumed for an influence diagram, meaning that a decision node and its parents are parents to all subsequent decisions. Finally, each utility node $U_j \in \mathbf{U}$ is associated with a utility function that depends only on the parents $pa(U_j)$ of $U_j$.

---
[*]This work was supported in part by the Science Foundation Ireland under grant no. 08/PI/I1912

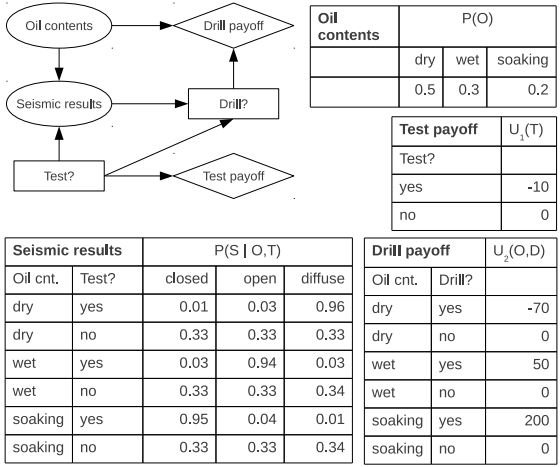

| Oil contents | P(O) | | |
|---|---|---|---|
| | dry | wet | soaking |
| | 0.5 | 0.3 | 0.2 |

| Test payoff | $U_1(T)$ |
|---|---|
| Test? | |
| yes | -10 |
| no | 0 |

| Seismic results | | P(S \| O,T) | | |
|---|---|---|---|---|
| Oil cnt. | Test? | closed | open | diffuse |
| dry | yes | 0.01 | 0.03 | 0.96 |
| dry | no | 0.33 | 0.33 | 0.33 |
| wet | yes | 0.03 | 0.94 | 0.03 |
| wet | no | 0.33 | 0.33 | 0.34 |
| soaking | yes | 0.95 | 0.04 | 0.01 |
| soaking | no | 0.33 | 0.33 | 0.34 |

| Drill payoff | | $U_2(O,D)$ |
|---|---|---|
| Oil cnt. | Drill? | |
| dry | yes | -70 |
| dry | no | 0 |
| wet | yes | 50 |
| wet | no | 0 |
| soaking | yes | 200 |
| soaking | no | 0 |

Figure 1: The oil wildcatter influence diagram.

The decision variables in an influence diagram are typically assumed to be temporally ordered. Let $D_1, D_2, ..., D_m$ be the order in which the decisions are to be made. The chance variables can be partitioned into a collection of disjoint sets $\mathbf{I}_0, \mathbf{I}_1, \ldots, \mathbf{I}_m$. For each $k$, where $0 < k < m$, $\mathbf{I}_k$ is the set of chance variables that are observed between $D_k$ and $D_{k+1}$. $\mathbf{I}_0$ is the set of initial evidence variables that are observed before the first $D_1$. $\mathbf{I}_m$ is the set of chance variables left unobserved when the last decision $D_m$ is made. This induces a partial order $\prec$ over $\mathbf{X} \cup \mathbf{D}$, as follows: $\mathbf{I}_0 \prec D_1 \prec \mathbf{I}_1 \prec \cdots \prec D_m \prec \mathbf{I}_m$ [5].

A *decision policy* (or strategy) for an influence diagram is a list of decision rules $\Delta = (\delta_1, \ldots, \delta_m)$ consisting of one rule for each decision variable. A *decision rule* for the decision $D_k \in \mathbf{D}$ is a mapping $\delta_k : \Omega_{pa(D_k)} \to \Omega_{D_k}$, where for a set $S \subseteq \mathbf{X} \cup \mathbf{D}$, $\Omega_S$ is the Cartesian product of the individual domains of the variables in $S$. Solving an influence diagram is to find the *optimal* decision policy that maximizes the expected utility. The maximum expected utility (MEU) is equal to:

$$\sum_{\mathbf{I}_0} \max_{D_1} \cdots \sum_{\mathbf{I}_{m-1}} \max_{D_m} \sum_{\mathbf{I}_m} \left( \prod_{i=1}^{n} P_i \times \sum_{j=1}^{r} U_j \right) \quad (1)$$

**Example 1** *For illustration, consider the influence diagram displayed in Figure 1 which is based on the classic oil wildcatter decision problem [9]. An oil wildcatter must decide either to* drill *or* not to drill *for oil at a specific site.*

*The wildcatter could do a* seismic test *that will help determine the* geological structure *of the site. The test results can show a* closed *reflection pattern (indication of significant oil), an* open *pattern (indication of some oil), or a* diffuse *pattern (almost no hope of oil). The probabilistic knowledge consists of the CPTs $P(O)$ and $P(S|O,T)$, while the utility function is the sum of $U_1(T)$ and $U_2(D,O)$. The optimal policy is to perform the seismic test and to drill only if the test results show an open or a closed pattern. The maximum expected utility of this policy is 42.75.*

**Variable Elimination** Several exact methods have been proposed over the past decades for solving influence diagrams using local computations [10, 13, 11, 5, 3, 8]. These methods adapted classical *variable elimination* techniques, which compute a type of marginalization over a combination of local functions, in order to handle the multiple types of information (probabilities and utilities), marginalization ($\sum$ and $\max$) and combination ($\times$ for probabilities, $+$ for utilities) involved in influence diagrams. Since the alternation of $\sum$ and $\max$ in Eq. 1 does not commute in general, it prevents the solution technique from eliminating variables in any ordering. Therefore, the computation dictated by Eq. 1 must be performed along a *legal elimination ordering* that respects $\prec$, namely the reverse of the elimination ordering is some extension of $\prec$ to a total order [5, 3].

## 3 FOUNDATIONS

Our approach towards a qualitative theory for influence diagrams is based on the qualitative decision theory proposed by Wilson [14]. Wilson's theory defines a set of abstract quantities called *extended reals*, denoted by $\mathbb{R}^*$, that are used to represent qualitative probabilities and utilities. Each extended real is a rational function $p/q$ where $p$ and $q$ are polynomials in $\epsilon$ with coefficients in the rationals, where $\epsilon$ is a very small but unknown quantity so that the extended reals can be used to represent information up to $\epsilon$ precision. For example, quantities such as $1-\epsilon$ and $\epsilon$ might be used for qualitative probabilities *likely* and *unlikely* respectively, and $\epsilon^{-1}$ for a high utility. These quantities can then be combined using standard arithmetic operations between polynomials for computing expected qualitative utilities. The resulting utilities are then compared among each other by means of a total order on $\mathbb{R}^*$ that is defined in [14].

### 3.1 ORDER-OF-MAGNITUDE CALCULUS

Rather than using extended reals explicitly, we adopt a simpler calculus that allows us to reason about the "order of magnitude" of the extended reals [14]. We start with the definition of an order-of-magnitude value that represents a qualitative probability or utility value.

DEFINITION 1 *An* order-of-magnitude value *is a pair*

$\langle \sigma, n \rangle$, where $\sigma \in \{+, -, \pm\}$ is called the sign and $n \in \mathbb{Z}$ is called the order of magnitude, respectively.

Intuitively, for each integer $n$ we have an element $\langle +, n \rangle$ meaning "of order $\epsilon^n$", and an element $\langle -, n \rangle$ meaning "of order $-\epsilon^n$". Moreover, if we add something of order $\epsilon^n$ to something of order $-\epsilon^n$ then the result can be of order $\pm \epsilon^m$, for any $m \geq n$. To ensure closure of the calculus under addition, we therefore add the element $\langle \pm, n \rangle$ representing this set of possibilities. In the following, we also define $\mathcal{O} = \{\langle \sigma, n \rangle \mid n \in \mathbb{Z}, \sigma \in \{+, -, \pm\}\} \cup \{\langle \pm, \infty \rangle\}$, $\mathcal{O}_{\pm} = \{\langle \pm, n \rangle \mid n \in \mathbb{Z} \cup \{\infty\}\}$ and $\mathcal{O}_{+} = \{\langle +, n \rangle \mid n \in \mathbb{Z} \cup \{\infty\}\}$. The element $\langle \pm, \infty \rangle$ will sometimes be written as 0, element $\langle +, 0 \rangle$ as 1, and element $\langle -, 0 \rangle$ as -1.

Standard arithmetic operations such as multiplication ($\times$) and addition ($+$) follow from the semantics of the order-of-magnitude values [14] and are defined next.

DEFINITION 2 (multiplication) *Let $a, b \in \mathcal{O}$ be such that $a = \langle \sigma, m \rangle$ and $b = \langle \tau, n \rangle$. We define $a \times b = \langle \sigma \otimes \tau, m + n \rangle$, where $\infty + n = n + \infty = \infty$ for $n \in \mathbb{Z} \cup \{\infty\}$ and $\otimes$ is the natural multiplication of signs, namely it is the commutative operation on $\{+, -, \pm\}$ such that $+ \otimes - = -$, $+ \otimes + = - \otimes - = +$, and $\forall \sigma \in \{+, -, \pm\}, \sigma \otimes \pm = \pm$.*

This multiplication is associative and commutative, and $\forall a \in \mathcal{O}, a \times 0 = 0$ and $a \times 1 = a$, respectively. Furthermore, for $b \in \mathcal{O} \setminus \mathcal{O}_{\pm}$, we define $b^{-1}$ to be the multiplicative inverse of $b$, namely $\langle \sigma, m \rangle^{-1} = \langle \sigma, -m \rangle$ for $\sigma \in \{+, -\}$. Given $a \in \mathcal{O}$, we define $a/b = a \times b^{-1}$.

DEFINITION 3 (addition) *Let $a, b \in \mathcal{O}$ be such that $a = \langle \sigma, m \rangle$ and $b = \langle \tau, n \rangle$. We define $a + b$ to be: (1) $\langle \sigma, m \rangle$ if $m < n$; (2) $\langle \tau, n \rangle$ if $m > n$; (3) $\langle \sigma \oplus \tau, m \rangle$ if $m = n$, where $+ \oplus + = +$, $- \oplus - = -$, and otherwise, $\sigma \oplus \tau = \pm$.*

Addition is associative and commutative, and $a + 0 = a$, $\forall a \in \mathcal{O}$. For $a, b \in \mathcal{O}$, let $-b = -1 \times b$ and $a - b = a + (-b)$. Clearly, we can write $-\langle \sigma, m \rangle = \langle -\sigma, m \rangle$, where $-(+) = -, -(-) = +$ and $-(\pm) = \pm$. We also have the distributivity: $\forall a, b, c \in \mathcal{O}, (a + b) \times c = a \times c + b \times c$.

### 3.2 ORDERING ON SETS OF ORDER-OF-MAGNITUDE VALUES

We will use the following ordering over the elements of $\mathcal{O}$, which is slightly stronger than that defined in [14].

DEFINITION 4 (ordering) *Let $a, b \in \mathcal{O}$ be such that $a = \langle \sigma, m \rangle$ and $b = \langle \tau, n \rangle$. We define the binary relation $\succcurlyeq$ on $\mathcal{O}$ by $a \succcurlyeq b$ if and only if either: (1) $\sigma = +$ and $\tau = +$ and $m \leq n$; or (2) $\sigma = +$ and $\tau = \pm$ and $m \leq n$; or (3) $\sigma = +$ and $\tau = -$; or (4) $\sigma = \pm$ and $\tau = -$ and $m \geq n$; or (5) $\sigma = -$ and $\tau = -$ and $m \geq n$.*

Given $a, b \in \mathcal{O}$, if $a \succcurlyeq b$ then we say that $a$ *dominates* $b$. For $A, B \subseteq \mathcal{O}$, we say that $A \succcurlyeq B$ if every element of $B$ is dominated by some element of $A$ (so that $A$ contains as least as large elements as $B$), namely if for all $b \in B$ there exists $a \in A$ with $a \succcurlyeq b$. As usual, we write $a \succ b$ if and only if $a \succcurlyeq b$ and it is not the case that $b \succcurlyeq a$. It is easy to see that $\succcurlyeq$ is a *partial order* on $\mathcal{O}$ and the following monotonicity property holds:

PROPOSITION 1 *Let $a, b, c \in \mathcal{O}$. If $a \succcurlyeq b$ then $a + c \succcurlyeq b + c$, and if $a \succcurlyeq b$ and $c \in \mathcal{O}_{+}$ then $a \times c \succcurlyeq b \times c$.*

Any finite set of order-of-magnitude values can therefore be represented by its maximal elements with respect to $\succcurlyeq$.

DEFINITION 5 (maximal set) *Given a finite set $A \subseteq \mathcal{O}$, we define the* maximal set *of $A$, denoted by $\max_{\succcurlyeq}(A)$, to be the set consisting of the undominated elements in $A$, namely $\max_{\succcurlyeq}(A) = \{a \in A \mid \nexists b \in A \text{ such that } b \succ a\}$.*

## 4 OPERATIONS ON SETS OF ORDER-OF-MAGNITUDE VALUES

We introduce now the main operations that can be performed over partially ordered finite sets of order-of-magnitude values. In particular, we extend the addition ($+$) and multiplication ($\times$) operations from singleton to sets of order-of-magnitude values as well as define a maximization operation over such sets.

### 4.1 ADDITION, MULTIPLICATION AND MAXIMIZATION

Given two finite sets $A, B \subseteq \mathcal{O}$ and $q \in \mathcal{O}_{+}$, we define the summation and multiplication operations as $A + B = \{a + b \mid a \in A, b \in B\}$ and $q \times A = \{q \times a \mid a \in A\}$, respectively. The maximization operation is defined by $\max(A, B) = \max_{\succcurlyeq}(A \cup B)$.

In order to use the order-of-magnitude calculus to define a qualitative version of influence diagrams we need to be sure that each of $+$, $\times$ and $\max$ is commutative and associative, and also to give sufficient conditions such that the following distributivity properties hold:

$\forall q, q_1, q_2 \in \mathcal{O}_{+}$ and $\forall A, B, C \subseteq \mathcal{O}$

D1 $q \times (A + B) = (q \times A) + (q \times B)$

D2 $(q_1 + q_2) \times A = (q_1 \times A) + (q_2 \times A)$

D3 $\max(A, B) + C = \max(A, C) + \max(B, C)$

It is easy to see that $+$, $\times$ and $\max$ are commutative and associative, and the distributivity properties (D1) and (D3) hold as well. Unfortunately, the distributivity property (D2) does not always hold for sets of order-of-magnitude values. To give a simple example, let $q_1 = \langle +, 2 \rangle$, $q_2 = \langle +, 3 \rangle$ and let $A = \{\langle \pm, 1 \rangle, \langle \pm, 4 \rangle\}$. Then, $(q_1 + q_2) \times A$ yields the set

$\{\langle \pm, 3\rangle, \langle \pm, 6\rangle\}$, whereas $(q_1 \times A) + (q_2 \times A)$ is equal to $\{\langle \pm, 3\rangle, \langle \pm, 4\rangle, \langle \pm, 6\rangle\}$. This property does however hold for convex sets, as we will show next.

## 4.2 CONVEX SETS AND CONVEX CLOSURE

Based on Definition 5, every element of a finite set $A \subseteq \mathcal{O}$ is dominated by some maximal element in $A$. We can therefore define an equivalence relation between finite sets of order-of-magnitude values, as follows.

DEFINITION 6 (**relation** $\approx$) *Given two finite sets $A, B \subseteq \mathcal{O}$, we say that $A$ is $\approx$-equivalent with $B$, denoted by $A \approx B$, if and only if $A \succcurlyeq B$ and $B \succcurlyeq A$.*

Clearly, $\approx$ is an equivalence relation, namely it is reflexive, symmetric and transitive. We then have that:

PROPOSITION 2 *Let $A, B, C \subseteq \mathcal{O}$ be finite sets and let $q \in \mathcal{O}_+$. The following properties hold: (1) $A \approx B$ if and only if $\max_{\succcurlyeq}(A) = \max_{\succcurlyeq}(B)$; (2) if $A \approx B$ then $A + C \approx B + C$ and $q \times A \approx q \times B$.*

We introduce next the notions of convex sets and convex closure of sets of order-of-magnitude values.

DEFINITION 7 *A set $A \subseteq \mathcal{O}$ is said to be* convex *if $\forall q_1, q_2 \in \mathcal{O}_+$ with $q_1 + q_2 = 1$, and $\forall a, b \in A$, we have that $(q_1 \times a) + (q_2 \times b) \in A$. The* convex closure $\mathcal{C}(A)$ *of a set $A \subseteq \mathcal{O}$ is defined to consist of every element of the form $\sum_{i=1}^{k}(q_i \times a_i)$, where $k$ is an arbitrary natural number, each $a_i \in A$, each $q_i \in \mathcal{O}_+$ and $\sum_{i=1}^{k} q_i = 1$.*

Consider two elements $\langle \sigma, m\rangle$ and $\langle \tau, n\rangle$ in $\mathcal{O}$, where we can assume without loss of generality that $m \leq n$. Any convex combination of these two elements is of the form $\langle \theta, l\rangle$ where $l \in [m, n]$ and if $l < n$ then $\theta = \sigma$; if $l = n$ then $\theta = \sigma \oplus \tau$ or $\theta = \tau$. This implies that the convex combination of a finite number of non-zero elements is finite (since every element $a$ in the convex combination has its order restricted to be within a finite range), and so, in particular can be represented by its maximal set. In fact, this property holds even if we allow the zero element $\langle \pm, \infty\rangle$. We can define now the following equivalence relation:

DEFINITION 8 (**relation** $\equiv$) *Given the finite sets $A, B \subseteq \mathcal{O}$, we say that $A$ is $\equiv$-equivalent with $B$, denoted by $A \equiv B$, if and only if $\mathcal{C}(A) \approx \mathcal{C}(B)$.*

Therefore, two sets of order-of-magnitude values are considered equivalent if, for every convex combination of elements of one, there is a convex combination of elements of the other which is at least as good.

PROPOSITION 3 *Let $A, B, C \subseteq \mathcal{O}$ be finite sets and let $q \in \mathcal{O}_+$. The following properties hold: (1) $A \equiv B$ if and only if $\max_{\succcurlyeq}(\mathcal{C}(A)) = \max_{\succcurlyeq}(\mathcal{C}(B))$; (2) if $A \equiv B$ then $A + C \equiv B + C$, $q \times A \equiv q \times B$, and $A \cup C \equiv B \cup C$.*

We can show now that any finite subset of $\mathcal{O}$ is in fact $\equiv$-equivalent with a set of order-of-magnitude values containing one or two elements, namely:

THEOREM 1 *Let $A$ be any finite subset of $\mathcal{O}$. Then either $A \equiv \{a\}$ for some $a \in \mathcal{O}$, or $\exists\, m, n \in \mathbb{Z}$ with $m < n$ and $\sigma \in \{+, -, \pm\}$ such that $A \equiv \{\langle \pm, m\rangle, \langle \sigma, n\rangle\}$.*

## 4.3 OPERATIONS ON EQUIVALENT SETS OF ORDER-OF-MAGNITUDE VALUES

Theorem 1 allows us to efficiently perform the required operations (ie, summation, multiplication and maximization) on sets of order-of-magnitude values. We assume that the subsets $\mathcal{O}$ are either singleton sets or are of the form $\{\langle \pm, m\rangle, \langle \sigma, n\rangle\}$, where $m < n$. We need to ensure that the outputs are of this form as well. For a given $a \in \mathcal{O}$, we use the notation $\sigma(a)$ and $\hat{a}$ to denote the sign and the order of magnitude of $a$, respectively,

**Multiplication** Given $A \subseteq \mathcal{O}$ of the required form, and $q \in \mathcal{O}_+$, we need to generate a set $A'$ that is $\equiv$-equivalent with $q \times A$. Write $q$ as $\langle +, l\rangle$. If $A = \{\langle \sigma, m\rangle\}$ then $q \times A$ is just equal to the singleton set $\{\langle \sigma, l+m\rangle\}$. Otherwise, $A$ is of the form $\{\langle \pm, m\rangle, \langle \sigma, n\rangle\}$, where $m < n$. Then $q \times A$ equals $\{\langle \pm, l+m\rangle, \langle \sigma, l+n\rangle\}$, which is of the required form, since $l + m < l + n$.

**Maximization** Given the sets $A_1, A_2, \ldots, A_k \subseteq \mathcal{O}$, each of them having the required form, we want to compute a set $A'$ that is $\equiv$-equivalent to $\max(A_1, \ldots, A_k)$. Let $A = A_1 \cup \cdots \cup A_k$ and, for $\sigma \in \{+, -, \pm\}$, we define $m_\sigma$ and $n_\sigma$ as follows: if there exists no element $a \in A$ with $\sigma(a) = \sigma$ then we say that $m_\sigma$ and $n_\sigma$ are both undefined; otherwise we have that $m_\sigma = \min\{l : \langle \sigma, l\rangle \in A\}$ and $n_\sigma = \max\{l : \langle \sigma, l\rangle \in A\}$, respectively. The set $A'$ is computed as follows: (1) if $m_+$ and $m_\pm$ are both undefined (there are only negative elements) then $A' = \{\langle -, n_-\rangle\}$; (2) if $m_+$ is defined and either $m_+ \leq m_\pm$ or $m_\pm$ is undefined then $A' = \{\langle +, m_+\rangle\}$; (3) if $m_+ > m_\pm$ (and both are defined) then $A' = \{\langle \pm, m_\pm\rangle, \langle +, m_+\rangle\}$; (4) if $m_+$ is undefined (no positive elements) and either $n_\pm \geq n_-$ or $n_-$ is undefined then $A' = \{\langle \pm, m_\pm\rangle, \langle \pm, n_\pm\rangle\}$ ; and (5) if $m_+$ is undefined (there are no positive elements) and $n_\pm < n_-$ then $A' = \{\langle \pm, m_\pm\rangle, \langle -, n_-\rangle\}$.

**Summation** Given the sets $A_1, A_2, \ldots, A_k \subseteq \mathcal{O}$ of required form as before, we want to compute a set $A'$ that is $\equiv$-equivalent to $(A_1 + \cdots + A_k)$. We can write $A_i$ as $\{a_i, b_i\}$ where if $a_i \neq b_i$ then $\sigma(a_i) = \pm$ and $\hat{a}_i < \hat{b}_i$. Then, $(A_1 + \cdots + A_k) \equiv \{a, b\}$ where $a = a_1 + \cdots + a_k$ and $b = b_1 + \cdots + b_k$. We can write $b$ more explicitly as $\langle \sigma(b), \hat{b}\rangle$ where $\hat{b} = \min(\hat{b}_1, \ldots, \hat{b}_k)$, and $\sigma(b) = +$ if and only if all $b_i$ with minimum $\hat{b}_i$ have $\sigma(b_i) = +$; else $\sigma(b) = -$ if all $b_i$ with minimum $\hat{b}_i$ have $\sigma(b_i) = -$; else $\sigma(b) = \pm$. Similarly for $a$. If $\sigma(a) \neq \pm$ then $\{a, b\}$ reduces to a singleton because $a = b$.

**Example 2** *Consider the sets $A_1 = \{\langle\pm, 3\rangle, \langle\pm, 4\rangle\}$ and $A_2 = \{\langle\pm, 3\rangle, \langle\pm, 6\rangle\}$. To generate $A' \equiv \max(A_1, A_2)$, we first compute $m_\pm = 3$ and $n_\pm = 6$, and then we have that $A' = \{\langle\pm, 3\rangle, \langle\pm, 6\rangle\}$ which corresponds to the extreme points of the input sets. Similarly, we can compute the set $A'' \equiv (A_1 + A_2)$ as $\{\langle\pm, 3\rangle, \langle\pm, 4\rangle\}$.*

### 4.4 DISTRIBUTIVITY PROPERTIES REVISITED

In summary, we can show now that all three distributivity properties hold with respect to the $\equiv$-equivalence relation between finite sets of order-of-magnitude values.

**THEOREM 2** $\forall q, q_1, q_2 \in \mathcal{O}_+$ and $\forall A, B, C \subseteq \mathcal{O}$ finite sets we have that: (D1) $q \times (A + B) \equiv (q \times A) + (q \times B)$; and (D2) $(q_1 + q_2) \times A \equiv (q_1 \times A) + (q_2 \times A)$; and (D3) $\max(A, B) + C \equiv \max(A, C) + \max(B, C)$.

## 5 ORDER-OF-MAGNITUDE INFLUENCE DIAGRAMS

In this section, we introduce a new qualitative version of the influence diagram model based on an order-of-magnitude representation of the probabilities and utilities.

### 5.1 THE QUALITATIVE DECISION MODEL

An *order-of-magnitude influence diagram* (OOM-ID) is a *qualitative* counterpart of the standard influence diagram graphical model. The graphical structure of an OOM-ID is identical to that of a standard ID, namely it is a directed acyclic graph containing *chance nodes* (circles) for the random discrete variables $\mathbf{X}$, *decision nodes* (rectangles) for the decision variables $\mathbf{D}$, and *utility nodes* (diamonds) for the local utility functions $\mathbf{U}$ of the decision maker. The directed arcs in the OOM-ID represent the same dependencies between the variables as in the standard model. Each chance node $X_i \in \mathbf{X}$ is associated with a conditional probability distribution $P_i^o$ that maps every configuration of its scope to a positive order-of-magnitude probability value, namely $P_i^o : \Omega_{X_i \cup pa(X_i)} \to \mathcal{O}_+$. The utility functions $U_j^o \in \mathbf{U}$ represent partially ordered preferences which are expressed by finite sets of order-of-magnitude values, namely $U_j^o : \Omega_{Q_j} \to 2^\mathcal{O}$, where $Q_j$ is the scope of $U_j$.

Solving an order-of-magnitude influence diagram is to find the *optimal policy* $\Delta = (\delta_1, \ldots, \delta_m)$ that maximizes the order-of-magnitude expected utility $\prod_{i=1}^n P_i^o \times \sum_{j=1}^r U_j^o$. We define the *optimal policies set* of an order-of-magnitude influence diagram to be the set of all policies having the same maximum order-of-magnitude expected utility.

### 5.2 AN EXAMPLE

Figure 2 displays the order-of-magnitude probability and utility functions of an OOM-ID corresponding to the oil wildcatter decision problem from Example 1. For our purpose, we used an extension of Spohn's mapping from the original probability distributions and utility functions to their corresponding order-of-magnitude approximation [12, 2]. Specifically, given a small positive $\epsilon < 1$, the order-of-magnitude approximation of a probability value $p \in (0, 1]$ is $\langle +, k \rangle$ such that $k \in \mathbb{Z}$ and $\epsilon^{k+1} < p \leq \epsilon^k$, while the order-of-magnitude approximation of a positive utility value $u > 0$ is $\langle +, -k \rangle$ such that $\epsilon^{-k} \leq u < \epsilon^{-(k+1)}$ (the case of negative utilities is symmetric). For example, if we consider $\epsilon = 0.1$ then the probability $P(S = closed | O = dry, T = yes) = 0.01$ is mapped to $\langle +, 2 \rangle$, while the utilities $U_2(O = dry, D = yes) = -70$ and $U_2(O = soaking, D = yes) = 200$ are mapped to $\langle -, -1 \rangle$ and $\langle +, -2 \rangle$, respectively.

| Seismic results | | P(S \| O,T) | | |
|---|---|---|---|---|
| Oil cnt. | Test? | closed | open | diffuse |
| dry | yes | (+,2) | (+,1) | (+,0) |
| dry | no | (+,0) | (+,0) | (+,0) |
| wet | yes | (+,1) | (+,0) | (+,1) |
| wet | no | (+,0) | (+,0) | (+,0) |
| soaking | yes | (+,0) | (+,1) | (+,2) |
| soaking | no | (+,0) | (+,0) | (+,0) |

| Oil contents | P(O) | | |
|---|---|---|---|
| | dry | wet | soaking |
| | (+,0) | (+,0) | (+,0) |

| Test payoff | $U_1(T)$ |
|---|---|
| Test? | |
| yes | {(-,-1)} |
| no | {(+,∞)} |

| Drill payoff | $U_2(O,D)$ | |
|---|---|---|
| Oil cnt. | Drill? | |
| dry | yes | {(-,-1)} |
| dry | no | {(+,∞)} |
| wet | yes | {(+,-1)} |
| wet | no | {(+,∞)} |
| soaking | yes | {(+,-2)} |
| soaking | no | {(+,∞)} |

Figure 2: Order-of-magnitude probability and utility functions corresponding to the oil wildcatter influence diagram.

Table 1: Optimal policies sets for order-of-magnitude influence diagrams corresponding to the oil wildcatter problem.

| decision rule | | OOM-ID | | |
|---|---|---|---|---|
| | | $\epsilon = 0.1$ | $\epsilon = 0.01$ | $\epsilon = 0.001$ |
| Test? | | {yes,no} | {yes, no} | {yes, no} |
| Drill? | S=closed, T=yes | yes | yes | {yes, no} |
| | S=open, T=yes | yes | yes | {yes, no} |
| | S=diffuse, T=yes | no | {yes, no} | {yes, no} |
| | S=closed, T=no | yes | yes | {yes, no} |
| | S=open, T=no | yes | yes | {yes, no} |
| | S=diffuse, T=no | yes | yes | {yes, no} |
| order-of-magnitude MEU | | $\langle +, -1 \rangle$ | $\langle +, 0 \rangle$ | $\{\langle \pm, 0\rangle, \langle +, \infty\rangle\}$ |

Table 1 shows the optimal policies sets (including the maximum order-of-magnitude expected utility) obtained for the order-of-magnitude influence diagrams corresponding to $\epsilon \in \{0.1, 0.01, 0.001\}$. When $\epsilon = 0.1$, we can see that there are two optimal policies having the same maximum order-of-magnitude expected utility, namely $\Delta_1$ (for $T = yes$) and $\Delta_2$ (for $T = no$). Therefore, if the seismic test is performed ($T = yes$) then drilling is to be done only if the test results show an open or closed pattern. Otherwise ($T = no$), the wildcatter will drill regardless of the test results. Ties like these at the decision variables are expected given that the order-of-magnitude probabilities and utilities represent abstractions of the real values. The expected utilities of $\Delta_1$ and $\Delta_2$ in the original influence diagram are

42.75 and 20.00, respectively.

When $\epsilon = 0.01$, we also see that both drilling options are equally possible if the seismic test is performed and the test results show a diffuse pattern. In this case, there are four optimal policies having the same maximum order-of-magnitude expected utility. Finally, when $\epsilon = 0.001$, we can see that all decision options are possible and the corresponding optimal policies set contains 128 policies. The explanation is that the order-of-magnitude influence diagram contains in this case only trivial order-of-magnitude values such as $\langle +, 0 \rangle$, $\langle -, 0 \rangle$ and $\langle +, \infty \rangle$, respectively.

### 5.3 VARIABLE ELIMINATION

Theorem 2 ensures the soundness and correctness of a variable elimination procedure using the summation (+), multiplication ($\times$) and maximization (max) operations over partially ordered sets of order-of-magnitude values, for solving order-of-magnitude influence diagrams.

Therefore, a variable elimination algorithm that computes the optimal policy of an order-of-magnitude influence diagram (and also the maximum order-of-magnitude expected utility) is described by Algorithm 1. The algorithm, called ELIM-OOM-ID, is based on Dechter's *bucket elimination* framework for standard influence diagrams [3] and uses a bucket structure constructed along a *legal elimination ordering* $o = Y_1, \ldots Y_t$ of the variables in $\mathbf{X} \cup \mathbf{D}$. The bucket data-structure, called *buckets*, associates each bucket with a single variable. The bucket of $Y_p$ contains all input probability and utility functions whose highest variable is $Y_p$.

The algorithm processes each bucket, top-down from the last to the first, by a variable elimination procedure that computes new probability (denoted by $\lambda$) and utility (denoted by $\theta$) components which are then placed in corresponding lower buckets (lines 1–11). The $\lambda^p$ of a chance bucket is generated by multiplying all probability components and eliminating by summation the bucket variable. The $\theta^p$ of a chance bucket is computed as the average utility of the bucket, normalized by the bucket's compiled $\lambda^p$. For a decision variable, we compute the $\lambda^p$ and $\theta^p$ components in a similar manner and eliminate the bucket variable by maximization. In this case, the product of the probability components in the bucket is a constant when viewed as a function of the bucket's decision variable [5, 15] and therefore, the compiled $\lambda^p$ is a constant as well.

In the second, bottom-up phase, the algorithm computes an optimal policy. The decision buckets are processed in reverse order, from the first variable to the last. Each decision rule is generated by taking the argument of the maximization operator applied over the combination of the probability and utility components in the respective bucket, for each configuration of the variables in the bucket's scope (ie, the union of the scopes of all functions in that bucket minus the

**Algorithm 1**: ELIM-OOM-ID

**Data**: An OOM-ID $\langle \mathbf{X}, \mathbf{D}, \mathbf{U}, G \rangle$, bucket structure along a legal elimination ordering of the variables $o$
**Result**: An optimal policy $\Delta$
// top-down phase
1 **for** $p = t$ *downto 1* **do**
2     let $\Lambda^p = \{\lambda_1, ..., \lambda_j\}$ and $\Theta^p = \{\theta_1, ..., \theta_k\}$ be the probability and utility components in $buckets[p]$
3     **if** $Y_p$ *is a chance variable* **then**
4        $\lambda^p \leftarrow \sum_{Y_p} \prod_{i=1}^{j} \lambda_i$
5        $\theta^p \leftarrow (\lambda^p)^{-1} \times \sum_{Y_p}((\prod_{i=1}^{j} \lambda_i) \times (\sum_{j=1}^{k} \theta_j))$
6     **else if** $Y_p$ *is a decision variable* **then**
7        **if** $\Lambda^p = \emptyset$ **then** $\theta^p \leftarrow \max_{Y_p} \sum_{j=1}^{k} \theta_j$
8        **else**
9           $\lambda^p \leftarrow \max_{Y_p} \prod_{i=1}^{j} \lambda_i$
10          $\theta^p \leftarrow \max_{Y_p}((\prod_{i=1}^{j} \lambda_i) \times (\sum_{j=1}^{k} \theta_j))$
11    place each $\lambda^p$ and $\theta^p$ in the bucket of the highest-index variable in its scope
// bottom-up phase
12 **for** $p = 1$ *to t* **do**
13     **if** $Y_p$ *is a decision variable* **then**
14        $\delta_p \leftarrow \arg\max_{Y_p}((\prod_{i=1}^{j} \lambda_i) \times (\sum_{j=1}^{k} \theta_j))$
15        $\Delta \leftarrow \Delta \cup \delta_p$
16 **return** $\Delta$

bucket variable $Y_p$).

THEOREM **3 (complexity)** *Given an OOM-ID with $n$ variables, algorithm ELIM-OOM-ID is time and space $O(n \cdot k^{w_o^*})$, where $w_o^*$ is the treewidth of the legal elimination ordering $o$ and $k$ bounds the domain size of the variables.*

## 6 EXPERIMENTS

In this section, we evaluate empirically the quality of the decision policies obtained for order-of-magnitude influence diagrams. All experiments were carried out on a 2.4GHz quad-core processor with 8GB of RAM.

**Methodology** We experimented with random influence diagrams described by the parameters $\langle n_c, n_d, k, p, r, a \rangle$, where $n_c$ is the number of chance variables, $n_d$ is the number of decision variables, $k$ is the maximum domain size, $p$ is the number of parents in the graph for each variable, $r$ is the number of root nodes and $a$ is the arity of the utility functions. The structure of the influence diagram is created by randomly picking $n_c + n_d - r$ variables out of $n_c + n_d$ and, for each, selecting $p$ parents from their preceding variables, relative to some ordering, whilst ensuring that the decision variables are connected by a directed path. A single utility node with $a$ parents picked randomly from the chance and decision nodes is then added to the graph.

We generated two classes of random problems with parameters $\langle n, 5, 2, 2, 5, 5 \rangle$ and having either positive utilities only or mixed (positive and negative) utilities. They are denoted by $P : \langle n, 5, 2, 2, 5, 5 \rangle$ and $M : \langle n, 5, 2, 2, 5, 5 \rangle$, respectively. In each case, 75% of the chance nodes were assigned extreme CPTs which were populated with numbers drawn uniformly at random between $10^{-5}$ and $10^{-4}$, whilst ensuring that the table is normalized. The remaining CPTs were randomly filled using a uniform distribution between 0 and 1. For class $P$, the utilities are of the form $10^u$, where $u$ is an integer uniformly distributed between 0 and 5. For class $M$, the utilities are of the form $+10^u$ or $-10^u$, where $u$ is between 0 and 5, as before, and we have an equal number of positive and negative utility values. Each influence diagram instance was then converted into a corresponding order-of-magnitude influence diagram using the mapping of the probabilities and utilities described in Section 5.2, for some $\epsilon < 1$. Intuitively, the smaller $\epsilon$ is, the coarser the order-of-magnitude approximation of the exact probability and utility values (ie, more information is lost).

**Measures of Performance** To measure how close the decision policies derived from the optimal policy set of an order-of-magnitude influence diagram are to the optimal policy of the corresponding standard influence diagram, we use two relative errors, defined as follows. Let $\mathcal{I}$ be an influence diagram and let $\mathcal{I}_\epsilon$ be the corresponding order-of-magnitude approximation, for some $\epsilon$ value. We sample $s$ different policies, uniformly at random, from the optimal policies set of $\mathcal{I}_\epsilon$, and for each sampled policy we compute its expected utility in $\mathcal{I}$. Let $\Delta_{med}$ be a policy corresponding to the median expected utility $v_{med}$ amongst the samples. We define the relative error $\eta_{med} = |(v - v_{med})/v|$, where $v$ is the maximum expected utility of the optimal policy in $\mathcal{I}$. Similarly, we define $\eta_{max} = |(v - v_{max})/v|$, where $\Delta_{max}$ is the best policy having the highest expected utility $v_{max}$ amongst the samples.

**Results** Figure 3 displays the distribution of the relative errors $\eta_{med}$ (top) and $\eta_{max}$ (bottom) obtained on order-of-magnitude influence diagrams derived from class $P$ (ie, positive utilities), as a function of the problem size (given by the number of variables), for $\epsilon \in \{0.5, 0.05, 0.005\}$. Each data point and corresponding error bar represents the $25^{th}$, median and $75^{th}$ percentiles obtained over 30 random problem instances generated for the respective problem size. We can see that $\eta_{med}$ is the smallest (less than 10%) for $\epsilon = 0.5$. However, as $\epsilon$ decreases, the loss of information due to the order-of-magnitude abstraction increases and the corresponding relative errors $\eta_{med}$ increase significantly. Notice that the best policy $\Delta_{max}$ derived from the order-of-magnitude influence diagram was almost identical to that of the corresponding standard influence diagram, for all $\epsilon$ (ie, the error $\eta_{max}$ is virtually zero).

Figure 4 shows the distribution of $\eta_{med}$ (top) and $\eta_{max}$ (bottom) obtained on order-of-magnitude influence diagrams from class $M$ (ie, mixed utilities). The pattern of the results is similar to that from the previous case. However, in this case, the errors span over two or three orders of magnitude, especially for $\epsilon = 0.05$ and $0.005$. This is because the sampled policy space includes policies which are quite different from each other and, although they have the same maximum order-of-magnitude expected utility, their expected utility in the corresponding standard influence diagram is significantly different. For this reason, we looked in more detail at the distribution of the expected utility values of 100 policies sampled uniformly at random from the optimal policies set of a class $M$ OOM-ID instance with 45 variables, for $\epsilon \in \{0.5, 0.05, 0.005\}$. As expected, we observed that the smallest sample variance is obtained for $\epsilon = 0.5$. For $\epsilon = 0.05$ and $\epsilon = 0.005$, the samples are spread out even more from the mean, and the variance of the expected utility is significantly larger. This explains the large variations of the relative errors $\eta_{med}$ and $\eta_{max}$, especially for smaller $\epsilon$ values (eg, $\epsilon = 0.05$ and $\epsilon = 0.005$).

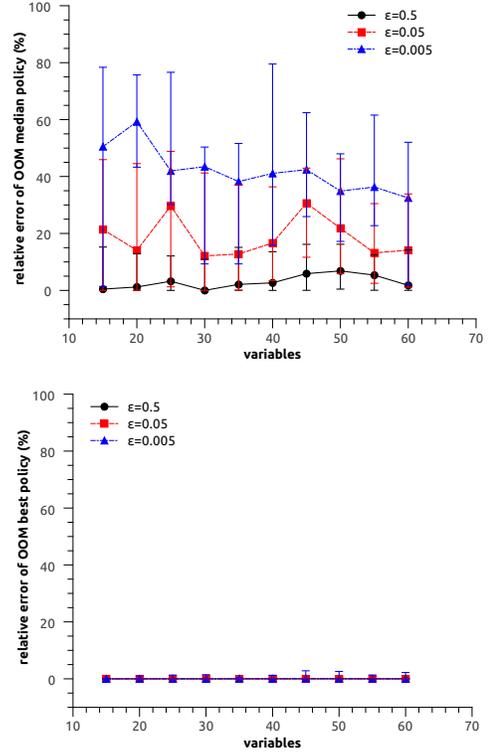

Figure 3: Results for class $P$ influence diagrams. We show the distribution of the relative errors $\eta_{med}$ (top) and $\eta_{max}$ (bottom) for $\epsilon \in \{0.5, 0.05, 0.005\}$. # of samples $s = 100$.

## 7 RELATED WORK

Several extensions of the standard influence diagram model have been proposed in recent years to deal with imprecise probabilistic and utility information. Garcia and Sabbadin [4] introduced possibilistic influence diagrams to model and solve decision making problems under qualitative uncertainty in the framework of possibilistic theory. Pralet *et al* [8] considered a generalized influence diagram system which allows for some qualitative uncertainty formalisms to be used together with totally ordered utility values only. Lopez and Rodriguez [7] proposed an influence diagram model based on random fuzzy variables to represent imprecise information. Kikuti *et al* [6] allow credal sets of probabilities to represent imprecise probabilistic information and focus on precise utility. The work that is closest to ours is that by Bonet and Pearl [1] who consider qualitative MDPs and POMDPs based also on an order-of-magnitude approximation of probabilities and totally ordered utilities.

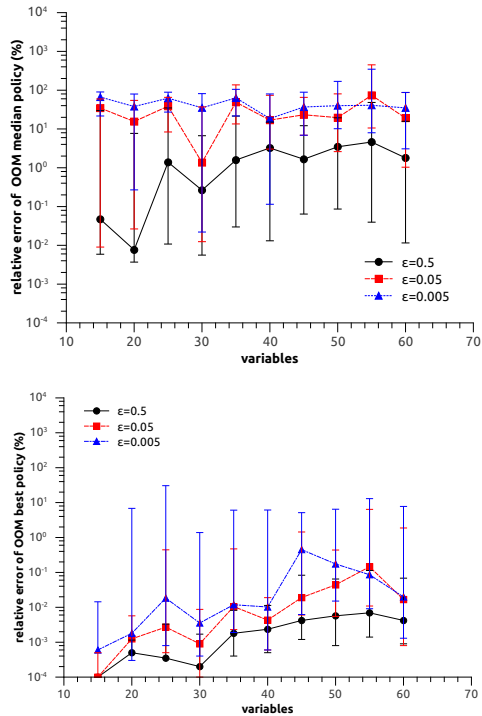

Figure 4: Results for class $M$ influence diagrams. We show the distribution of the relative errors $\eta_{med}$ (top) and $\eta_{max}$ (bottom) for $\epsilon \in \{0.5, 0.05, 0.005\}$. # of samples $s = 100$.

## 8 CONCLUSION

The paper presents a qualitative influence diagram formalism that allows reasoning with imprecise probabilities and partially ordered imprecise utility values. Our proposed order-of-magnitude influence diagram model is based on an order-of-magnitude representation of the probabilities and utilities. We also described a dedicated variable elimination algorithm that performs efficient operations on partially ordered sets of utilities for solving this model.

We considered a straightforward variable elimination algorithm. One way to improve it is to efficiently exploit constraints (zero values of the uncertainty and utility functions), building, for instance, on work by [8]. We also plan to investigate other formalisms for reasoning with imprecise information, such as interval-valued utilities and multi-attribute utility allowing trade-offs.